\documentclass[11pt,a4paper]{article}
\usepackage{authblk}
\usepackage[hyperref]{acl2020}
\usepackage{times}
\usepackage{latexsym}

\usepackage{microtype}

\usepackage{graphicx}
\usepackage{tikz}
\usepackage{subcaption}
\usepackage{adjustbox}
\usepackage{multirow}
\usepackage{booktabs}
\usepackage[linewidth=1pt]{mdframed}

\usepackage{amsmath}

\aclfinalcopy 

\title{QUACKIE: A NLP Classification Task With Ground Truth Explanations}

\author[1,2]{Yves Rychener}
\author[2]{Xavier Renard}
\author[3]{Djamé Seddah}
\author[1]{Pascal Frossard}
\author[2,4,5]{Marcin Detyniecki}
\affil[1]{EPFL, Lausanne, Switzerland}
\affil[2]{AXA, Paris, France}
\affil[3]{Inria, Paris, France}
\affil[4]{Sorbonne Université, Paris, France}
\affil[5]{Polish Academy of Science, Warsaw, Poland}

\date{}

\begin{document}
\maketitle
\begin{abstract}
NLP Interpretability aims to increase trust in model predictions. This makes evaluating interpretability approaches a pressing issue.
There are multiple datasets for evaluating NLP interpretability, but their dependence on human provided ground truths raises questions about their unbiasedness. In this work, we take a different approach and formulate a specific classification task by diverting question-answering datasets. For this custom classification task, the interpretability ground-truth arises directly from the definition of the classification problem. We use this method to propose a benchmark and lay the groundwork for future research in NLP interpretability by evaluating a wide range of current state of the art methods.
\end{abstract}

\section{Introduction}
\label{sec:intro}
In recent years, we have seen an increase in the performance of NLP models in a variety of tasks. It has, to a large extent, been fuelled by the usage of progressively complex neural network architectures \cite{devlin2019bert, brown2020GPT3}. While this allows to model increasingly complex input-output relations, it is virtually impossible to understand the reasoning behind a prediction. To address this issue, algorithms have been proposed to explain the reasoning of models in the field of interpretability. While there has been a number of interpretability approaches published which are applicable to NLP,  \cite{simonyan2013saliency, smilkov2017smoothgrad, sundararajan2017integratedgrads, ribeiro2016LIME, lundberg2017SHAP}, defining what constitutes a good explanation and appropriate evaluation methods are largely unanswered problems.

This work considers the case of post-hoc interpretability for classification, where the reasoning of an existing classification model is explained. However, the evaluation technique could be adopted to also test the explanation performance of inherently interpretable classification models.

Interpretability and explanation are very subjective terms, with different definitions for different needs~\citep{Liao2020}. 
However, to our knowledge, the field agrees that \textit{faithfulness} is a desired property of a good interpretability approach. Faithfulness means that an explanation well replicates the actual reasoning of the model to explain. However, since the actual reasoning of a black-box model is unknown, the task of evaluating faithfulness is difficult. Many works proposing interpretability algorithms have thus resorted (often without explicitly acknowledging) to \textit{plausibility} for evaluating faithfulness, evaluating how convincing an explanation is to a human. Recently, \newcite{jacovi2020eval} pointed out the drawbacks of this approach, coming to the conclusion that many current evaluation approaches in the NLP interpretability literature may be biased towards matching human's intuition instead of explaining the black-box model.

In this work, we propose a novel evaluation approach for faithfulness. It is based on the observation that a good interpretability algorithm should be able to well distinguish between \textit{important} and \textit{unimportant} parts of the text.
Our evaluation methodology can compare any interpretability algorithm assigning importance for parts of the inputs. By building a classification task on top of the NLP question-answering problem, we are able to construct a ground truth label for \textit{important} and \textit{unimportant} parts of the input text with respect to the classification problem. We further show how these labels can be leveraged for more detailed metrics. The methodology is applied to two question-answering datasets in order to propose a new benchmark, QUACKIE, for interpretability evaluation.

This work is structured as follows: Section~\ref{sec:target} motivates our approach. In Section~\ref{sec:QA}, we present the classification task, construction of ground truths and the metrics used. This is followed by applying the methods to propose the QUACKIE benchmark and analysing its properties in Section~\ref{sec:QA:Task-Analysis}. Section~\ref{sec:Algos} presents a selection of interpretability algorithms, which aims to reflect the current state of post-hoc Black-Box and White-Box interpretability research. The algorithms are compared in Section~\ref{sec:Experiment} using QUACKIE. In Section~\ref{sec:related_work}, we discuss the work and its results.\footnote{The code for this work is available on GitHub and designed such that researchers can easily apply it to assess their own interpretability algorithms. Appendix~\ref{app:custom_interpreters} gives an overview of the code structure.}
The main contributions of this work are the construction of interpretability ground truth values which do not need human annotation and its application to compare state of the art post-hoc interpretability methods.

\section{Evaluating Discriminative Power of Interpretability Methods}
\label{sec:target}

No matter the explanation needs or the interpretability definition, a key property of a good interpretability method is its fidelity to the model to explain. To evaluate this property, we propose an approach to assess interpretability methods' fidelity. We provide an evaluation method for a universal requirement of NLP interpretability: The ability to distinguish between relevant and irrelevant text segments for the classification. In certain applications, an even finer separation may be useful, however the desideratum stated here builds a common ground.

Some works exist to assess interpretability methods: their quality is evaluated using datasets built on human generated ground-truths on top of common classification tasks~\cite{deyoung2019eraser}. As stated by \newcite{jacovi2020eval}, such human generated gold labels are built on assumptions about what a model \textit{should} do according to humans. This can lead to problems, as the actual evaluation of the interpretability method shifts from creating a faithful explanation for the reasoning of the model, to giving an "explanation" that best agrees with human reasoning for the text-prediction pair. The result is that high scoring interpretability methods could suffer from the disastrous problem of giving convincing explanations which do not match the reasoning of the classification model. As noted by \newcite{jacovi2020eval}, a \textit{plausible} but \textit{unfaithful} explanation may be the worst-case scenario, promoting trust in a prediction even if it is not warranted.

Further, since they do not provide a model to interpret, the benchmark results may become polluted: If wrong model predictions are not excluded from the benchmark, advances in classification accuracy lead to advances in interpretation performance. This is because we cannot expect the rationale for a faulty prediction to be the same as for a correct one, which is used as ground truth. If this problem is addressed by excluding samples with wrong model predictions, then different samples are excluded for different classifiers, which leads to different versions of the dataset being used for different interpretability methods, calling comparability into question.

We use a different approach to combat the issue of ground truth interpretability labels. Instead of annotating ground truth interpretability labels for an existing classification task, we develop a classification task for which ground-truth importance labels arise directly from the problem definition. If the model is successful in classification, it must base its prediction on these rationales. We also address the problem of benchmark pollution by providing trained classification models, which achieve good classification performance and ensure a level playing field.

In the next section, we will provide an description of how this classification task is created as well as explain the metrics used.

\section{Approach}
\label{sec:QA}
We will now present our interpretability evaluation approach for NLP. It is based on the question-answering (QA) task, where an answer span to a given question has to be identified in a text. Note that while we will later use this approach on two specific datasets for QUACKIE, the approach is general and applicable for any QA dataset, provided it contains the answers' location in the text and single sentence reasoning can be assumed.
\subsection{Ground Truth Construction}
\label{sec:QA:Approach_gt}
\subsubsection{The Question-Answering Task}
We base our evaluation procedure on an extractive question answering task, where the answer needs not necessarily be in the context.
A model is trained to extract the answer $\mathcal{A}$ to a question $\mathcal{Q}$ from a given textual context $\mathcal{C}$. However, it is also possible that there is no answer in the context, a fact which models also have to identify. In other words, the QA model $f$ is trained to predict
\begin{equation}
\label{eq:qa:approach:f}
f(\mathcal{Q}, \mathcal{C}) =\left\{
\begin{array}{ll}
\mathcal{A} & \mathcal{A} \subseteq \mathcal{C} \\
\emptyset & \, \textrm{otherwise} \\
\end{array}
\right..
\end{equation}
Due to the nature of the extractive question-answering task, the ground truth answer is not only supplied as text, but also its span in the context is given in most datasets (Examples Figure~\ref{fig:qa:approach:squadsample}). This can be leveraged to identify the sentence containing the answer to the question.

\begin{figure}
\begin{mdframed}
\textsf{\small{
    \textbf{Context}\\
    The further decline of Byzantine state-of-affairs paved the road to a third attack in {\textbf{\color{blue}1185}}, when a large Norman army invaded Dyrrachium, owing to the betrayal of high Byzantine officials. Some time later, Dyrrachium—one of the most important naval bases of {\textbf{\color{red}the Adriatic}}—fell again to Byzantine hands.\\
    \\
    \textbf{Some Questions}\\
    When did the Normans attack Dyrrachium? {\textbf{\color{blue}1185}}\\
    Where was Dyrrachium located? {\textbf{\color{red}the Adriatic}}\\
    Who attacked Dyrrachium in the 11th century? {\textbf{\color{gray}No Answer}}}}
\end{mdframed}
    \caption{An sample of questions-answers from SQuAD 2.0~\cite{rajpurkar2018squad2}, \textit{Normans from validation set}}
    \label{fig:qa:approach:squadsample}
\end{figure}

\subsubsection{Classification Task Based on Question-Answering}

The answer to the question and its surroundings must certainly be some of the most important parts for correctly detecting the answer and consequently the presence of an answer in the context.
This suggests that a model trained on predicting if an answer is present in the context must place its focus on the sentence containing the answer. In other words, the most important sentence for knowing that an answer is in the context is the sentence containing the answer.

\begin{figure}
\vskip 0.2in
\begin{center}
\begin{tikzpicture}[x=0.75pt,y=0.75pt,yscale=-1,xscale=1,scale=0.6, every node/.style={scale=0.6}]

\draw  [fill={rgb, 255:red, 0; green, 0; blue, 0 }  ,fill opacity=0.2 ] (260.25,131) -- (340.25,131) -- (340.25,171) -- (260.25,171) -- cycle ;

\draw    (300.25,79) -- (300.25,118) ;
\draw [shift={(300.25,120)}, rotate = 270] [color={rgb, 255:red, 0; green, 0; blue, 0 }  ][line width=0.75]    (10.93,-3.29) .. controls (6.95,-1.4) and (3.31,-0.3) .. (0,0) .. controls (3.31,0.3) and (6.95,1.4) .. (10.93,3.29)   ;
\draw    (192,151) -- (250,151) ;
\draw [shift={(252,151)}, rotate = 180] [color={rgb, 255:red, 0; green, 0; blue, 0 }  ][line width=0.75]    (10.93,-3.29) .. controls (6.95,-1.4) and (3.31,-0.3) .. (0,0) .. controls (3.31,0.3) and (6.95,1.4) .. (10.93,3.29)   ;
\draw    (350,151) -- (388,151) ;
\draw [shift={(390,151)}, rotate = 180] [color={rgb, 255:red, 0; green, 0; blue, 0 }  ][line width=0.75]    (10.93,-3.29) .. controls (6.95,-1.4) and (3.31,-0.3) .. (0,0) .. controls (3.31,0.3) and (6.95,1.4) .. (10.93,3.29)   ;
\draw    (480,151) -- (518,151) ;
\draw [shift={(520,151)}, rotate = 180] [color={rgb, 255:red, 0; green, 0; blue, 0 }  ][line width=0.75]    (10.93,-3.29) .. controls (6.95,-1.4) and (3.31,-0.3) .. (0,0) .. controls (3.31,0.3) and (6.95,1.4) .. (10.93,3.29)   ;
\draw  [color={rgb, 255:red, 74; green, 144; blue, 226 }  ,draw opacity=1 ][line width=1.5]  (220,50) -- (500,50) -- (500,180) -- (220,180) -- cycle ;

\draw (261.25,61) node [anchor=north west][inner sep=0.75pt]   [align=left] {Question ($\mathcal{Q}$)};
\draw (111,142.5) node [anchor=north west][inner sep=0.75pt]   [align=left] {Context ($\mathcal{C}$)};
\draw (278.75,132) node [anchor=north west][inner sep=0.75pt]   [align=left] {\begin{minipage}[lt]{30.504375000000003pt}\setlength\topsep{0pt}
\begin{center}
Q\&A\\Model
\end{center}

\end{minipage}};
\draw (399,142.5) node [anchor=north west][inner sep=0.75pt]   [align=left] {Answer ($\mathcal{A}$)};
\draw (526,142.5) node [anchor=north west][inner sep=0.75pt]  [color={rgb, 255:red, 74; green, 144; blue, 226 }  ,opacity=1 ] [align=left] {$\mathcal{A}\neq0?$};

\end{tikzpicture}
\caption{Illustration of the SQuAD experiment: Black/Grey: The Question-Answering Problem with the pretrained Model $f$, Blue: Our proposed classification task for evaluating NLP interpretability methods ($g$).}
\label{fig:qa:approach:squad_architecture}
\end{center}
\vskip -0.2in
\end{figure}
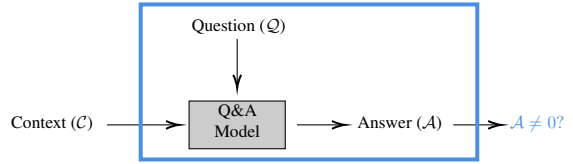

While this seems trivial, this intuition can be leveraged to construct a classification task with ground truth interpretability labels.
We build a classification task $g$ on top of an existing question-answering dataset, where models have to predict if the answer to the question is in the context or not. 
For this, we view the question as fixed for each question-context pair, thus building a classification task with only the context as input. An already trained QA model is used, which means no training is needed:
\begin{equation}
\label{eq:g}
    g(\mathcal{C}) = \textbf{1}\{f(\mathcal{Q}, \mathcal{C}) \neq \emptyset \}
\end{equation}
where $f$ is according to the definition in Equation~\ref{eq:qa:approach:f} and $\textbf{1}\{\cdot\}$ is the indicator function. A visual depiction of this can be seen in Figure~\ref{fig:qa:approach:squad_architecture}.

Then, interpretability is performed to identify the most important sentence for samples where the answer is in the context. This prediction is compared to the ground truth: the sentence containing the answer. 

Note that only the correctly classified samples are used (\textit{True Positive}). This is because, looking at the confusion matrix, \textit{True Negative} and \textit{False Positive} cannot be considered since no ground truth can be derived from the underlying dataset (since no answer exists). We also exclude \textit{False Negative} samples, since the obtained ground truth is for a different prediction than the one being made.

One may think that this evaluation procedure is in fact a case of \textit{What should the model do?} evaluation.
However, in the original SQuAD paper \cite{rajpurkar2016squad}, authors find that in the majority of cases (86.4\%), single sentence reasoning is sufficient to find the answer.
Further, employing an erasure based view, removing the sentence with the answer will remove the answer, resulting in the answer no longer being present in the text. The question is thus no longer answerable, leading to a change in prediction. This implies that the sentence containing the answer must be the most important one for a positive prediction. A quantitative test of this is given in Section~\ref{sec:QA:Task-Analysis} for the datasets used in QUACKIE.

\subsubsection{Sentence as Ground Truth}
\label{sec:approach_sentence}
Many interpretability methods provide word-based explanations. It would thus seem natural to directly take the answer itself as ground truth and not the whole sentence containing it. However, we argue that the answer itself is not an appropriate interpretability ground truth. Consider the question \textit{"What is the capital of France?"} and a context containing the sentence \textit{"[...] The capital of France is Paris. [...]"}. 
We cannot expect a classifier to know that the question is answerable with just \textit{"Paris"} as the rationale, since this requires external knowledge. However, it would seem that a language model, which only argues based on the text, can know that the answer to the question is in the context by looking at the rationale \textit{"capital of France is"}: The reasoning is not based on the answer itself, but on its surrounding. 
To account for this, we use sentences as ground truth. Interpretability methods producing explanations of other granularity (e.g. words) should aggregate them to the sentence level. For the baselines examined in our benchmark, it is not clear how the token scores should be combined into a sentence score, arguments can be made for both taking the maximum score and summing the scores in each sentence. We thus test both methods in our benchmark.

\subsection{Metrics}
\label{sec:QA:Metrics}
Interpretability algorithms may attribute importance to inputs in a variety of ways: By attributing \textbf{scores} to input tokens, by \textbf{ranking} tokens according to their importance or by \textbf{selecting} \textit{important} vs \textit{unimportant} tokens. Of course, scores can be used to produce a ranking, and the ranking can be used to perform selection. However, we believe it is important to use metrics that accommodate for the nuances in attributions, which is why we propose three metrics for the three levels of information: selection, ranking and scoring. All of them are aggregated across samples by reporting their mean.

\subsubsection{IoU for Evaluating Selections}
Given a ground truth sentence and one or more selected sentences, we use the IoU (Intersection over Union) between them as score. If the interpreter selects only one sentence, then the IoU is either 0 or 1, thus after aggregation reflecting the \textit{accuracy of detection}. 
Given binary vectors of selection and ground truth ($s$ and $g$) respectively, the score is:
$$
IoU = \frac{\sum (s\land g)}{\sum(s \lor g)},
$$
where $\sum$ signifies summing over all entries, with True being 1 and False being 0. This metric reflects how well the ground truth is detected. 

\subsubsection{HPD for Evaluating Rankings}
If a ranking of sentences is provided by the interpreter, then we may leverage this for more detailed scoring: Even if the ground truth is not detected as the most important sentence, it should be among the highest in the ranking. To measure this property, we propose HPD (Highest Precision for Detection). Instead of selecting the most important sentence as prediction, we select the K-most important ones, where K is the rank of the ground truth. For this selection, the precision is reported. Given a ranking of sentences in decreasing importance $r$ and ground truth sentence $GT$:

\begin{align*}
    HPD = &\min_k \frac{1}{k} \\
    &\text{s.t  } GT \in r_1^k
\end{align*}
This metric measures how highly ranked the ground truth is.

\subsubsection{SNR for Evaluating Scores}
When scores are provided, we cannot simply compare them to a common target, since scores from different interpretability methods mean different things: Saliency scores reflect the aggregated local gradient, while Integrated Gradient scores reflect the aggregated effects on a the transition from a baseline to the text sample.
However, since we know that only one sentence is important, while the rest is not, we would like this sentence to have a significantly higher score than the rest. This can be measured with an adapted SNR (signal to noise ratio), which we define as follows: Denote $s$ the scores of the sentences and $GT$ the index of the ground truth sentence. Denote $s_{GT}$ the score of the ground truth sentence and $s_{\backslash GT}$ the scores of all other sentences. Using $mean$ and $std$, the mean and standard deviation, we define:
$$
    SNR = \frac{(s_{GT}-mean(s_{\backslash GT}))^2}{std(s_{\backslash GT})^2}
$$
This metric measures the selectivity of the interpretation. Note that while the other metrics directly reflect performance, this metric could in theory be optimized for, by decreasing the standard deviation of sentences predicted as non-important in post-processing. Nonetheless, we believe that interesting conclusions can be drawn from this metric.

Since IoU and HPD both measure how well the explanation reflects model behaviour, while SNR reflects selectiveness of the scores, we see IoU and HPD as primary metrics, which SNR as an additional, secondary metric.

\section{QUACKIE}
\label{sec:QA:Task-Analysis}
\begin{table}[!htb]
\vskip 0.15in
\begin{center}
\begin{small}
\begin{sc}
\resizebox{\linewidth}{!}{\begin{tabular}{l|rrrrr}
\toprule
Model   & SQuAD & New Wiki & NYT & Reddit & Amazon \\
\midrule
QA      & 5066 & 6857 & 8317 & 5121 & 5910 \\
Classif & 5597 & 7494 & 9444 & 7882 & 8321 \\
\bottomrule
\end{tabular}}
\subcaption*{Number of Samples Used in each Dataset}
\bigskip
\resizebox{\linewidth}{!}{\begin{tabular}{l|rrrrr}
\toprule
Model   & SQuAD & New Wiki & NYT & Reddit & Amazon \\
\midrule
QA & 5.32 & 5.06 & 5.86 & 8.75 & 7.64 \\
Classif & 5.32& 5.05 & 5.92 & 8.81 & 7.74 \\
\bottomrule
\end{tabular}}
\subcaption*{Average Number of Sentences}

\bigskip
\resizebox{\linewidth}{!}{\begin{tabular}{l|rrrrr}
\toprule
Model   & SQuAD & New Wiki & NYT & Reddit & Amazon \\
\midrule
QA      & 0.79& 0.86& 0.91 & \textbf{0.56} & 0.75 \\
Classif & 0.82 & 0.97 & 0.96 & 0.88 & 0.88 \\
\bottomrule
\end{tabular}}
\subcaption*{Classifier Accuracy}
\end{sc}
\end{small}
\end{center}
\caption{Descriptive Statistics of Datasets Used}
\label{table:datasets}
\vskip -0.1in
\end{table}

So far we have presented the methodology for extracting ground-truths for classification interpretability evaluation from QA datasets and introduced the metrics. We apply this methodology to two datasets in order to propose a new benchmark for interpretability evaluation; QUACKIE\footnote{QUACKIE stands for \textbf{QU}estion and \textbf{A}nswering for \textbf{C}lassification tas\textbf{K} \textbf{I}nterpretability \textbf{E}valuation}. We further provide two pretrained models to interpret. Finally, we perform a dataset analysis as well as examining the accuracy of the ground truth.
\subsection{Datasets}
We use two datasets in our benchmark:

\textbf{SQuAD 2.0}~\cite{rajpurkar2018squad2} is a question answering dataset based on a selection of Wikipedia articles. It has the exact task explained in Section~\ref{sec:QA:Approach_gt}.

\textbf{SQuADShifts}~\cite{miller2020squadshifts} gives 4 new test sets from different domains for SQuAD: New Wiki (other Wikipedia Contexts), NYT (New York Times Articles), Reddit (Reddit Posts) and Amazon (Amazon Reviews). The aim is to provide different semantic contexts for SQuAD. However, in this dataset the answer is always in the context. Since we use pretrained models from SQuAD 2.0, this is not a problem.

\subsection{Classification Models}
We use two pretrained models which are available in the HuggingFace Transformers library~\citep{wolf-etal-2020-transformers} and were trained on SQuAD 2.0.

The first model is a RoBERTa~\cite{liu2019roberta} based classifier\footnote{\url{https://huggingface.co/a-ware/roberta-large-squad-classification}} which was directly trained on the classification problem $g(\cdot)$ from Equation~\ref{eq:g}. It is hereafter named "Classif" model.

The second model is an ALBERT~\cite{lan2019albert} question answering model~\footnote{\url{https://huggingface.co/twmkn9/albert-base-v2-squad2}}, trained on the task $f(\cdot, \cdot)$ from Equation~\ref{eq:qa:approach:f}. From the output scores, we perform a softmax to get probabilities for start- and end-tokens. From these, we can deduct the probability of no answer being present (invalid span or answer in question or CLS token). It is hereafter named "QA" model.

\subsection{Descriptive Statistics}
As mentioned before, we only use True Positive predictions for interpretability analysis. Further, some samples are discarded because their tokenization is too long for the models (Question + Context $>$ 512 tokens). This results in a different number of samples usable for the different classifiers and datasets. To get a sense of the datasets used, we give the number of samples and average number of sentences for each dataset (after filtering). We also report the accuracy of the models over the whole dataset (positive and negative samples). Since the SQuADShifts datasets only contain positives, the recall (fraction of correctly predicted positive samples) is equal to the accuracy. For SQuAD, the recall is 0.91 and 0.82 for the Classif and QA models respectively. See Table~\ref{table:datasets} for the full results. With the exception of the QA model on Reddit (marked in \textbf{bold}), accuracy and recall are consistently high, suggesting that model predictions can be trusted, making them good candidates for interpretation benchmarking.

\subsection{Ground Truth Verification}
In the original SQuAD paper~\cite{rajpurkar2016squad}, the authors perform an analysis into the type of reasoning required for finding the answer and find that in the majority of cases (86.4\%), only single sentence reasoning is necessary. This motivated us to use the sentence containing the answer as ground-truth. We will however now evaluate this ground truth with two experiments:

The experiments used are similar to the metrics \textit{comprehensiveness} and \textit{sufficiency} used by \newcite{deyoung2019eraser}. In the first experiment, we remove the sentence containing the ground truth and observe the change in probability output of the model. In the second experiment, we remove the rest of the text and only keep the sentence containing the answer. We again observe the change in probability output. 
The results are given in Table~\ref{table:gt_eval}, the probability output without removing anything can be seen in Appendix~\ref{app:prob_output}. Note that for the classification model, predictions are very discriminative, which results in the change in probability being either 0 or 1. Nonetheless, we observe for both datasets that removing the ground truth results in a much bigger decrease in probability than removing all the rest. This suggests that our ground-truth labels are accurate, since the ground truth rationales are necessary (comprehensiveness) and sufficient (sufficiency) for prediction.

\begin{table*}[!htb]
    \begin{minipage}{.5\linewidth}
      \centering
        \resizebox{0.9\linewidth}{!}{\begin{tabular}{l|rrrrr}
            \toprule
            Model   & SQuAD & New Wiki & NYT & Reddit & Amazon \\
            \midrule
            QA      & -0.70 & -0.67 & -0.73 & -0.63 & -0.62\\
            Classif & -0.62 & -0.60 & -0.68 & -0.54 & -0.56\\
            \bottomrule
        \end{tabular}}
    \caption*{Remove Ground Truth (Comprehensiveness)}
    \end{minipage}%
    \begin{minipage}{.5\linewidth}
      \centering
        \resizebox{0.9\linewidth}{!}{\begin{tabular}{l|rrrrr}
            \toprule
            Model   & SQuAD & New Wiki & NYT & Reddit & Amazon \\
            \midrule
            QA      & -0.04 & -0.04 & -0.05 & -0.04 & -0.05\\
            Classif & -0.03 & -0.03 & -0.04  & -0.18 & -0.11 \\
            \bottomrule
        \end{tabular}}
    \caption*{Only Keep Ground Truth (Sufficiency)}
    \end{minipage} 
    \caption{Average Change in Probability Output}
\label{table:gt_eval}
\end{table*}

\section{Baselines}
Having formulated our benchmark, we will now give a selection of interpretability algorithms which we compare in order to lay a foundation for the comparison of interpretability approaches with QUACKIE.
\subsection{Interpretability Algorithms}
\label{sec:Algos}
There is not an agreement on the problem formulation of NLP interpretability. This means that many approaches use different definitions, resulting in differing types of outputs and assumptions. As it was already stated in Section~\ref{sec:intro}, we consider algorithms that map parts of the inputs to their \textit{importances}, essentially creating a heatmap (sometimes referred to as the attribution problem). We split the considered interpretability algorithms into 2 groups, based on the assumptions of model access:
\begin{itemize}
    \item \textbf{White-Box} approaches assume complete access to the model and its internals, notably access to gradients.
    \item \textbf{Black-Box} approaches do not need access to model internals, the model is viewed as a black box. We do however allow access to probability outputs, a condition that is not always necessary.
\end{itemize}
We will now present the interpretability algorithms we considered. For a more detailed explanation of the approaches, refer to their original papers.~
\subsubsection{White-Box}
\label{sec:Algos:WB}

\textbf{Sailency Maps} \cite{simonyan2013saliency} report the gradient of the output with respect to each input token as the importance of the token. While saliency maps are a straight-forward solution to the attribution problem, they have been observed to create noisy attributions~\cite{smilkov2017smoothgrad}. The gradient is aggregated for each token by summing over all the dimensions. Saliency maps are the simplest attribution method based on gradients and often used as baseline for gradient-based methods.

\textbf{SmoothGrad/NoiseTunnel}~\cite{smilkov2017smoothgrad} aims to address the issue of noise by averaging the attributions from multiple similar inputs, which are created by adding small amounts of noise to the original input. The gradient is aggregated for each token by summing over all the dimensions. SmootGrad is a widely used approach, as its approach of \textit{removing noise by adding noise} has been shown to reduce noise in attributions, making it a viable candidate for benchmarking.

\textbf{Integrated Gradients} \cite{sundararajan2017integratedgrads} follow an axiomatic approach to the attribution problem by approximating the path integral of the gradients between a baseline reference and the input to explain. Multiplication by the inputs is used, the final scores are aggregated across token dimensions taking the sum. The axiomatic approach makes Integrated Gradients an interesting addition to our benchmark.

For all Black-Box methods, we use the probability of the \textit{answerable} class as target for gradient computation.
\subsubsection{Black-Box}
\label{sec:Algos:BB}

\textbf{LIME} \cite{ribeiro2016LIME} uses a weighted linear regression to locally approximate the decision function. A local dataset is created by randomly removing some words from the text. After fitting the weighted linear regression, the coefficients are used as word-attributions. We use LIME in our benchmark since it is widely used and a staple in black-box interpretability.

\textbf{SHAP} \cite{lundberg2017SHAP} follows the same approach as LIME and constructs a local linear approximation of the decision function. It utilizes methods from cooperative game-theory to derive a kernel for the weighted regression. Since SHAP provides guarantees under certain assumptions, it is interesting to look at it as a baseline.

\subsubsection{Baseline}
\label{sec:Algos:Baseline}
The algorithms are compared to a random baseline, which creates a random ordering of sentences, by attributing them with evenly spaced scores from $0$ to $n_{words}-1$.
\subsection{Results}
\label{sec:Experiment}
\begin{figure*}
    \centering
    \includegraphics[width=\linewidth, trim=5cm 2cm 5cm 2cm ]{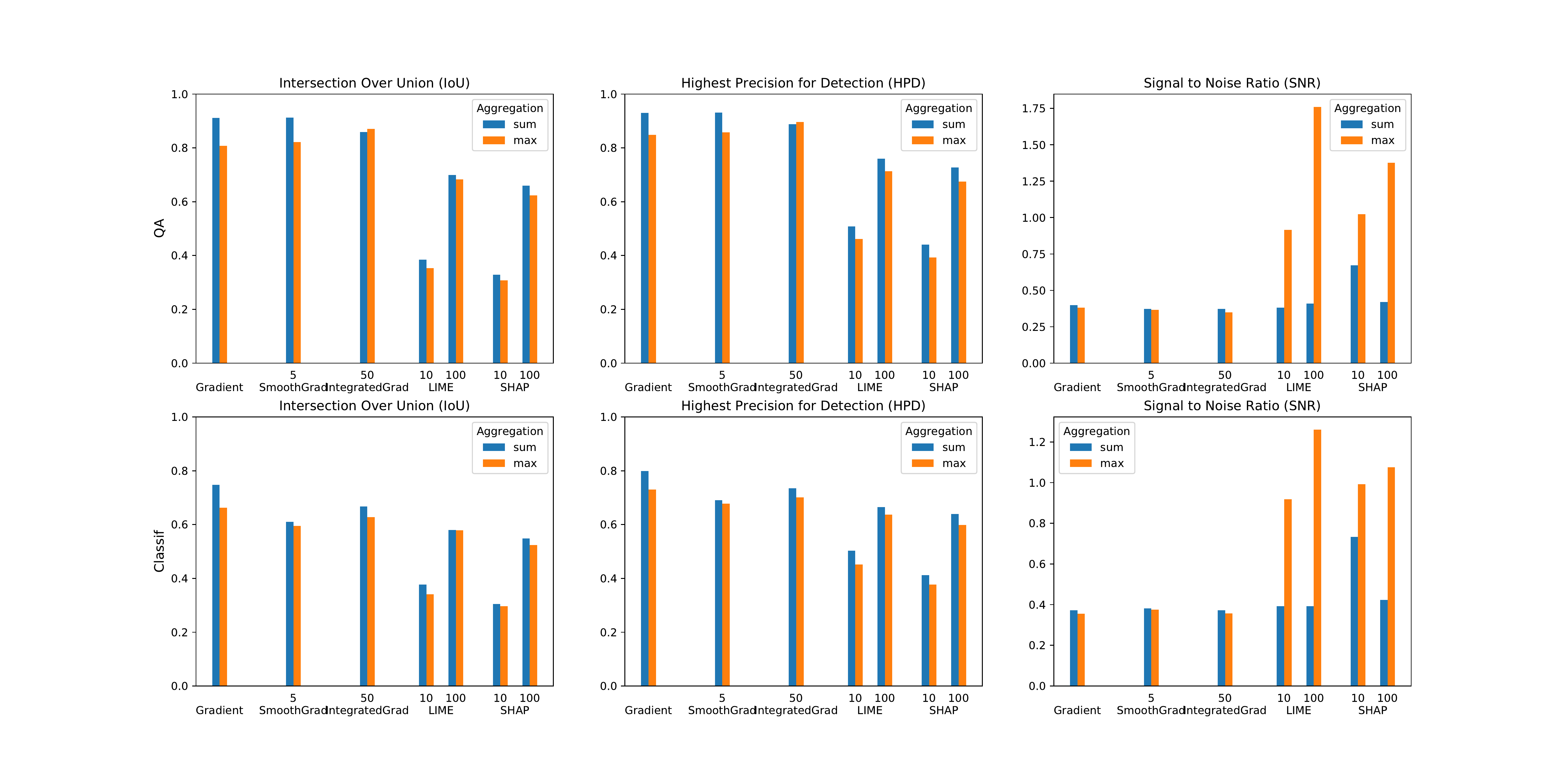}
    \caption{Benchmark Results on SQuAD 2.0}
    \label{fig:res:squad}
\end{figure*}

Figure~\ref{fig:res:squad} displays the results on the SQuAD 2.0 dataset in visual form. Complete Results in tabular form can be found in Appdendix~\ref{app:complete_results}.
We compare the interpretation approaches with different aggregation methods (sum and max, as stated in Section~\ref{sec:approach_sentence}).

First, we observe that model access can greatly improve the explanation accuracy, reflected by the consistently higher scores of Saliency and Smoothgrad. Further, the performance of Black-Box models LIME and SHAP increases significantly with the number of samples used. Evaluation with the default number of samples was not possible due to computational limitations: Evaluation with a neighborhood of 100 samples took around 10h for LIME on only SQuAD 2.0 for 1 classifier\footnote{Experiments were carried out on an AWS SageMaker instance with 1/2 Nvidia Tesla K80}, using the default value of 5000 samples seems infeasible with reasonable computational resources, as we expect it to take close to 200 days only for LIME on a single GPU instance. We further note, surprisingly, that SmoothGrad does not seem to bring much improvement over standard Saliency using gradients. We attribute this to the fact that noise reduction is already occurring by averaging the scores over token dimensions and over the sentences. 
Finally, note that both LIME and SHAP use LASSO regression as a surrogate, which results in the attribution of a score of $0$ to unimportant sentences, thus reducing the noise, which increases the SNR.

\section{Related Work and Discussion}
\label{sec:related_work}

Most work in evaluating rationales has been done by comparing rationales to human judgements~\cite{doshi2017towards, strout2019human} or by letting humans evaluate the explanations~\cite{ribeiro2016LIME, lundberg2017SHAP}. While this approach measures the plausibility of interpretations, it does not guarantee that explanations actually reflect model behaviour. \newcite{jacovi2020eval} performs an in-depth analysis on the state of current NLP interpretability evaluation approaches and desired future properties, notably the need to evaluate faithfulness, as plausible but unfaithful explanations are arguably the worst-case scenario. However, it would seem that all evaluation methodology based on human-annotated ground truth labels is biased towards plausibility by construction.
\newcite{deyoung2019eraser} address this problem by using the additional metrics \textit{comprehensiveness} and \textit{sufficiency}, which measure change in model prediction when removing the rationale and only keeping the rationale. Our approach also aims to address the bias towards plausibility, but instead of addressing the problem with metrics, we build a classification task for which interpretability ground truth labels arise directly without human annotation. To our knowledge, it is the first approach to propose unbiased ground truths for rationales.

The take-away from our benchmark results is two-fold. First, for practitioners, with the current state of the art we see that model access should be used if possible, as it is able to significantly increase the fidelity of explanations. However this access is not always possible, in particular with complex text processing pipelines. Thus, for researchers, as Black-Box models lag behind gradient-based methods by a substantial amount, especially with limited computational capacity (10 samples), it seems that there is still a large potential for advances in black-box interpretability.

Finally, we note that the results and descriptive statistics for the \textit{Reddit} and \textit{Amazon} datasets are quite different from the other datasets (Table~\ref{table:datasets}, \ref{table:gt_eval}). We hypothesize that this is because they contain user-generated content, which has different properties that standard text~\cite{eisenstein2013bad}.

\section{Conclusion}
\label{sec:Discussion}
In this work, we proposed a novel model-agnostic NLP interpretability evaluation methodology for text classification. Our methodology allows for a ground truth that arises directly from the definition of the classification task to be interpreted. Further, it avoids biases of current state of the art evaluation methodologies based on human's intuition. We used this methodology to develop a benchmark based on two different Question-Answering datasets with a variety of semantic contexts. 

We publish our code, data and models to facilitate testing of future methods, which will hopefully lead to the continuous comparison of interpretability approaches. By providing results on a range of currently available interpretability methods, we provide an easy way for researchers to compare their approach to the state of the art. The publication of these results can also aid NLP practitioners in selecting a suitable interpretability approach. We encourage researchers to submit their results\footnote{Researchers may make a pull request on GitHub with their results. Website: \url{https://axa-rev-research.github.io/quackie/}}, as it helps the community keep track of the advances in NLP interpretability.

\section*{Acknowledgements}
We would like to thank Travis McGuire (huggingface ID \textsc{twmkn9}) and A-Ware UG (huggingface ID \textsc{a-ware}) for publishing their pretrained models, which we use for our benchmark.
Djamé Seddah was party founded by the French National Research Agency via the ANR ParSiTi project (ANR-16-CE33-002).

\bibliography{anthology,acl2020}

\begin{thebibliography}{19}
\expandafter\ifx\csname natexlab\endcsname\relax\def\natexlab#1{#1}\fi

\bibitem[{Brown et~al.(2020)Brown, Mann, Ryder, Subbiah, Kaplan, Dhariwal,
  Neelakantan, Shyam, Sastry, Askell et~al.}]{brown2020GPT3}
Tom~B Brown, Benjamin Mann, Nick Ryder, Melanie Subbiah, Jared Kaplan, Prafulla
  Dhariwal, Arvind Neelakantan, Pranav Shyam, Girish Sastry, Amanda Askell,
  et~al. 2020.
\newblock Language models are few-shot learners.
\newblock \emph{arXiv preprint arXiv:2005.14165}.

\bibitem[{Devlin et~al.(2019)Devlin, Chang, Lee, and
  Toutanova}]{devlin2019bert}
Jacob Devlin, Ming-Wei Chang, Kenton Lee, and Kristina Toutanova. 2019.
\newblock Bert: Pre-training of deep bidirectional transformers for language
  understanding.
\newblock In \emph{Proceedings of the 2019 Conference of the North American
  Chapter of the Association for Computational Linguistics: Human Language
  Technologies, Volume 1 (Long and Short Papers)}, pages 4171--4186.

\bibitem[{DeYoung et~al.(2019)DeYoung, Jain, Rajani, Lehman, Xiong, Socher, and
  Wallace}]{deyoung2019eraser}
Jay DeYoung, Sarthak Jain, Nazneen~Fatema Rajani, Eric Lehman, Caiming Xiong,
  Richard Socher, and Byron~C Wallace. 2019.
\newblock Eraser: A benchmark to evaluate rationalized nlp models.
\newblock \emph{arXiv preprint arXiv:1911.03429}.

\bibitem[{Doshi-Velez and Kim(2017)}]{doshi2017towards}
Finale Doshi-Velez and Been Kim. 2017.
\newblock Towards a rigorous science of interpretable machine learning.
\newblock \emph{arXiv preprint arXiv:1702.08608}.

\bibitem[{Eisenstein(2013)}]{eisenstein2013bad}
Jacob Eisenstein. 2013.
\newblock What to do about bad language on the internet.
\newblock In \emph{Proceedings of the 2013 conference of the North American
  Chapter of the association for computational linguistics: Human language
  technologies}, pages 359--369.

\bibitem[{Jacovi and Goldberg(2020)}]{jacovi2020eval}
Alon Jacovi and Yoav Goldberg. 2020.
\newblock Towards faithfully interpretable nlp systems: How should we define
  and evaluate faithfulness?
\newblock \emph{arXiv preprint arXiv:2004.03685}.

\bibitem[{Lan et~al.(2019)Lan, Chen, Goodman, Gimpel, Sharma, and
  Soricut}]{lan2019albert}
Zhenzhong Lan, Mingda Chen, Sebastian Goodman, Kevin Gimpel, Piyush Sharma, and
  Radu Soricut. 2019.
\newblock Albert: A lite bert for self-supervised learning of language
  representations.
\newblock \emph{arXiv preprint arXiv:1909.11942}.

\bibitem[{Liao et~al.(2020)Liao, Gruen, and Miller}]{Liao2020}
Q.~Vera Liao, Daniel Gruen, and Sarah Miller. 2020.
\newblock {Questioning the AI: Informing Design Practices for Explainable AI
  User Experiences}.
\newblock \emph{Conference on Human Factors in Computing Systems -
  Proceedings}.

\bibitem[{Liu et~al.(2019)Liu, Ott, Goyal, Du, Joshi, Chen, Levy, Lewis,
  Zettlemoyer, and Stoyanov}]{liu2019roberta}
Yinhan Liu, Myle Ott, Naman Goyal, Jingfei Du, Mandar Joshi, Danqi Chen, Omer
  Levy, Mike Lewis, Luke Zettlemoyer, and Veselin Stoyanov. 2019.
\newblock Roberta: A robustly optimized bert pretraining approach.
\newblock \emph{arXiv preprint arXiv:1907.11692}.

\bibitem[{Lundberg and Lee(2017)}]{lundberg2017SHAP}
Scott~M Lundberg and Su-In Lee. 2017.
\newblock A unified approach to interpreting model predictions.
\newblock In \emph{Advances in neural information processing systems}, pages
  4765--4774.

\bibitem[{Miller et~al.(2020)Miller, Krauth, Recht, and
  Schmidt}]{miller2020squadshifts}
John Miller, Karl Krauth, Benjamin Recht, and Ludwig Schmidt. 2020.
\newblock The effect of natural distribution shift on question answering
  models.
\newblock \emph{arXiv preprint arXiv:2004.14444}.

\bibitem[{Rajpurkar et~al.(2018)Rajpurkar, Jia, and
  Liang}]{rajpurkar2018squad2}
Pranav Rajpurkar, Robin Jia, and Percy Liang. 2018.
\newblock Know what you don’t know: Unanswerable questions for squad.
\newblock pages 784--789.

\bibitem[{Rajpurkar et~al.(2016)Rajpurkar, Zhang, Lopyrev, and
  Liang}]{rajpurkar2016squad}
Pranav Rajpurkar, Jian Zhang, Konstantin Lopyrev, and Percy Liang. 2016.
\newblock Squad: 100,000+ questions for machine comprehension of text.
\newblock In \emph{Proceedings of the 2016 Conference on Empirical Methods in
  Natural Language Processing}, pages 2383--2392.

\bibitem[{Ribeiro et~al.(2016)Ribeiro, Singh, and Guestrin}]{ribeiro2016LIME}
Marco~Tulio Ribeiro, Sameer Singh, and Carlos Guestrin. 2016.
\newblock " why should i trust you?" explaining the predictions of any
  classifier.
\newblock In \emph{Proceedings of the 22nd ACM SIGKDD international conference
  on knowledge discovery and data mining}, pages 1135--1144.

\bibitem[{Simonyan et~al.(2013)Simonyan, Vedaldi, and
  Zisserman}]{simonyan2013saliency}
Karen Simonyan, Andrea Vedaldi, and Andrew Zisserman. 2013.
\newblock Deep inside convolutional networks: Visualising image classification
  models and saliency maps.
\newblock \emph{arXiv preprint arXiv:1312.6034}.

\bibitem[{Smilkov et~al.(2017)Smilkov, Thorat, Kim, Vi{\'e}gas, and
  Wattenberg}]{smilkov2017smoothgrad}
Daniel Smilkov, Nikhil Thorat, Been Kim, Fernanda Vi{\'e}gas, and Martin
  Wattenberg. 2017.
\newblock Smoothgrad: removing noise by adding noise.
\newblock \emph{arXiv preprint arXiv:1706.03825}.

\bibitem[{Strout et~al.(2019)Strout, Zhang, and Mooney}]{strout2019human}
Julia Strout, Ye~Zhang, and Raymond~J Mooney. 2019.
\newblock Do human rationales improve machine explanations?
\newblock \emph{arXiv preprint arXiv:1905.13714}.

\bibitem[{Sundararajan et~al.(2017)Sundararajan, Taly, and
  Yan}]{sundararajan2017integratedgrads}
Mukund Sundararajan, Ankur Taly, and Qiqi Yan. 2017.
\newblock Axiomatic attribution for deep networks.
\newblock In \emph{International Conference on Machine Learning}, pages
  3319--3328.

\bibitem[{Wolf et~al.(2020)Wolf, Debut, Sanh, Chaumond, Delangue, Moi, Cistac,
  Rault, Louf, Funtowicz, Davison, Shleifer, von Platen, Ma, Jernite, Plu, Xu,
  Scao, Gugger, Drame, Lhoest, and Rush}]{wolf-etal-2020-transformers}
Thomas Wolf, Lysandre Debut, Victor Sanh, Julien Chaumond, Clement Delangue,
  Anthony Moi, Pierric Cistac, Tim Rault, Rémi Louf, Morgan Funtowicz, Joe
  Davison, Sam Shleifer, Patrick von Platen, Clara Ma, Yacine Jernite, Julien
  Plu, Canwen Xu, Teven~Le Scao, Sylvain Gugger, Mariama Drame, Quentin Lhoest,
  and Alexander~M. Rush. 2020.
\newblock \href {https://www.aclweb.org/anthology/2020.emnlp-demos.6}
  {Transformers: State-of-the-art natural language processing}.
\newblock In \emph{Proceedings of the 2020 Conference on Empirical Methods in
  Natural Language Processing: System Demonstrations}, pages 38--45, Online.
  Association for Computational Linguistics.

\end{thebibliography}
\bibliographystyle{acl_natbib}
\newpage
\appendix
\section{Testing Custom Interpretability algorithms}
\label{app:custom_interpreters}
\begin{figure}[hbt!]
    \centering
    \includegraphics[trim=2cm 6cm 17cm 1cm, clip, width=0.3\textwidth]{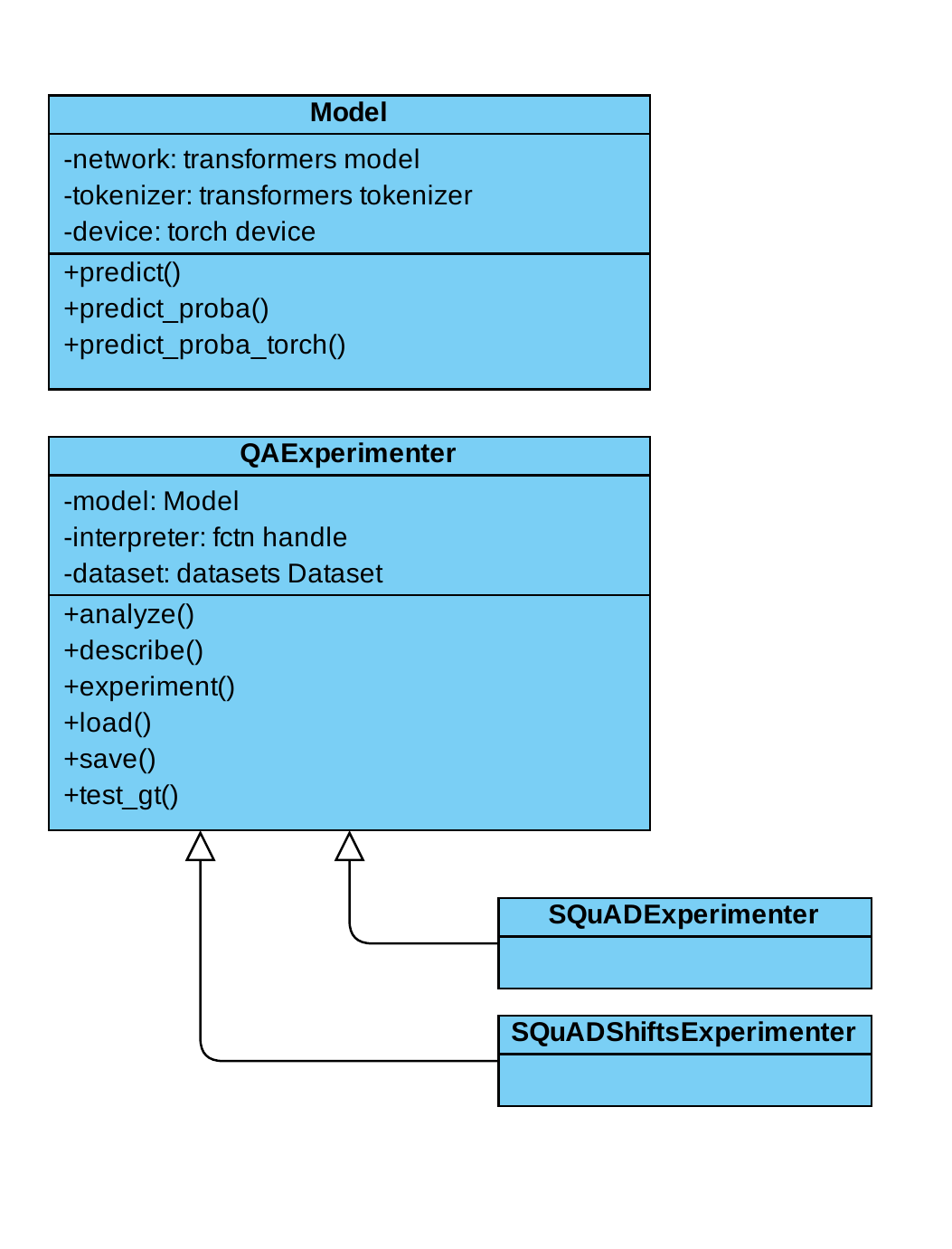}
    \caption{Class Diagram of the Experimentation Suite}
    \label{fig:app_custom:class_diagram}
\end{figure}

Figure~\ref{fig:app_custom:class_diagram} gives the UML class diagram of the experimentation framework. The central part of the framework is the \textit{QAExperimenter}. It can also be instantiated with \textit{SQuADExperimenter} or \textit{SQuADShiftsExperimenter}, which automatically load the corresponding NLP dataset. 

The user should provide the instantiated classification model (from \textit{Model\_QA} or \textit{Model\_Classification}) and interpreter function. The interpreter function receives the model, question and context as arguments. It should produce a numpy array of importances, of the same length as the number of sentences in the context (using \textit{nltk.tokenize.sent\_tokenize}). For information on how to use the model for other interpreters, one may refer to the implemented interpreters in the \textit{interpreters} folder. The file \textit{run\_experiment.py} can be used to run the experiment with custom interpreters, \textit{run\_sota.py} was used to produce the results published here. 

\section{Mean Probability Output}
\label{app:prob_output}
To complement the results from Table~\ref{table:gt_eval}, we give the average probability output with the complete context present in Table~\ref{tab:app:model_prediction}.

\begin{table}[!ht]
\vskip 0.15in
\begin{center}
\begin{small}
\begin{sc}
\resizebox{\linewidth}{!}{\begin{tabular}{l|rrrrr}
\toprule
Model   & SQuAD & New Wiki & NYT & Reddit & Amazon \\
\midrule
QA      & 0.94 & 0.95 & 0.94 & 0.88 & 0.89 \\
Classif & 1.00 & 1.00 & 1.00 & 1.00 & 1.00 \\
\bottomrule
\end{tabular}}
\end{sc}
\end{small}
\end{center}
\caption{Average Probability Output for Question-Context Pairs, round to two digits}
\label{tab:app:model_prediction}
\vskip -0.1in
\end{table}

\section{Complete Results in Tabular Form}
\label{app:complete_results}
Tables~\ref{tab:res:iou}, \ref{tab:res:hpd} and \ref{tab:res:snr} give the complete results on the SQuAD 2.0 and SQuADShifts datasets. The results on SQuADShifts are in line with the findings on SQuAD 2.0.

\begin{table*}[h]
    \centering
    \begin{adjustbox}{width=\textwidth}
    \begin{tabular}{lll|rr|rr|rr|rr|rr}
        &&&\multicolumn{2}{c|}{SQuAD}&\multicolumn{2}{c|}{New Wiki}&\multicolumn{2}{c|}{NYT}&\multicolumn{2}{c|}{Reddit}&\multicolumn{2}{c}{Amazon}\\
        Interpreter & Aggregation & Samples & Classif & QA& Classif & QA& Classif & QA& Classif & QA& Classif & QA\\
        \hline
        \hline
        \multirow{4}*{LIME\textsuperscript{\dag}} & \multirow{2}*{sum} & 10 & 37.70& 38.47 &40.72&41.40&40.22&41.98&31.35&35.34&32.69&35.27\\
                            &                    & 100 & 58.04 & 69.90&60.74&70.82&62.02&73.50&50.33&69.02&53.57&67.58\\
                            & \multirow{2}*{max} & 10 &34.06&35.36&36.98&37.77&36.19&36.43&26.52&28.89&29.80&32.12\\
                            &                    & 100 &57.86&68.30&59.65&68.43&61.57&72.00&48.22&65.23&53.09&66.24\\
        \hline
        \multirow{4}*{SHAP\textsuperscript{\dag}} & \multirow{2}*{sum} & 10 & 30.48 & 32.90&31.66&32.84&29.26&31.03&22.13&23.75&24.59&25.43\\
                            &                    & 100 & 54.85 & 65.92&57.53&65.79&56.38&67.70&49.35&65.02&54.03&67.68\\
                            & \multirow{2}*{max} & 10 &  29.69&30.81&30.72&31.68&28.32&30.00&21.17&22.58&22.72&23.84\\
                            &                    & 100 &52.45&62.34&54.56&63.18&53.19&64.78&45.79&60.03&49.54&63.35\\
        \hline
        \hline
        \multirow{2}*{Saliency} & sum & - & 74.74 & 91.12 &72.19&91.18&68.87&88.46&57.57&85.26&64.82&85.91\\
                                & max & - &66.27&80.79&65.04&80.78&58.95&76.07&48.41&77.33&59.52&79.70\\
        \hline
        \multirow{2}*{Integrated Gradients} & sum & 50 & 66.73& 85.93& 65.00& 85.93& 65.44& 85.20& 51.62& 79.73& 51.96& 78.21\\
                                            & max & 50 &62.73& 87.05& 60.70& 86.73& 61.63& 85.92& 50.24& 82.35& 49.09& 82.45\\
        \hline
        \multirow{2}*{SmoothGrad} & sum & 5 & 60.98&91.28 &60.29&90.56&60.25&88.29&50.32&84.51&52.34&84.40\\
                                  & max & 5 &59.48&82.16&61.45&82.38&56.93&78.33&45.95&77.72&53.03&78.26\\
        \hline
        \hline
        Random & - & - & 24.64&25.38 &26.86&27.39&24.53&24.36&16.51&16.09&18.71&19.17\\
    \end{tabular}
    \end{adjustbox}
    \caption{IoU Results
    (\dag Note that evaluation with the default number of samples was not possible due to computational limitations (100 samples already took 7h and 14h for Classif and QA model on SQuAD))}
    \label{tab:res:iou}
\end{table*}

\begin{table*}[h]
    \centering
    \begin{adjustbox}{width=\textwidth}
    \begin{tabular}{lll|rr|rr|rr|rr|rr}
        &&&\multicolumn{2}{c|}{SQuAD}&\multicolumn{2}{c|}{New Wiki}&\multicolumn{2}{c|}{NYT}&\multicolumn{2}{c|}{Reddit}&\multicolumn{2}{c}{Amazon}\\
        Interpreter & Aggregation & Samples & Classif & QA& Classif & QA& Classif & QA& Classif & QA& Classif & QA\\
        \hline
        \hline
        \multirow{4}*{LIME\textsuperscript{\dag}} & \multirow{2}*{sum} & 10 &50.29&50.83&53.32&53.76&51.62&53.12&39.99&43.56&42.31&44.64\\
                            &                    & 100 &66.50&75.98&68.93&76.93&69.17&78.58&56.60&72.97&60.12&72.25\\
                            & \multirow{2}*{max} & 10 &45.12&46.19&47.85&48.62&46.60&47.11&34.47&36.74&38.43&40.63\\
                            &                    & 100 &63.74&71.33&65.47&71.89&67.28&75.25&53.23&67.65&58.41&69.21\\
        \hline
        \multirow{4}*{SHAP\textsuperscript{\dag}} & \multirow{2}*{sum} & 10 &41.22&44.09&42.87&44.57&39.06&41.38&28.94&31.26&32.97&34.63\\
                            &                    & 100 &63.93&72.75&66.18&72.91&64.39&73.74&55.59&69.44&60.48&72.29\\
                            & \multirow{2}*{max} & 10 &37.74&39.28&39.29&40.68&36.40&38.54&27.30&29.13&30.24&32.05\\
                            &                    & 100 &59.85&67.47&61.80&68.35&60.80&69.98&51.19&63.41&55.32&66.86\\
        \hline
        \hline
        \multirow{2}*{Saliency} & sum & - &79.91&93.01&78.20&93.06&74.97&90.71&63.05&87.21&69.96&88.01\\
                                & max & - &72.99&84.83&72.40&84.81&66.86&80.94&55.10&80.32&65.31&82.74\\
        \hline
        \multirow{2}*{Integrated Gradients} & sum & 50 &73.52& 88.85& 72.39& 88.93& 71.99& 88.00& 57.84& 82.46& 58.93& 81.56\\
                                            & max & 50 &70.15& 89.67& 68.93& 89.48& 68.73& 88.52& 56.51& 84.70& 56.35& 85.10\\
        \hline
        \multirow{2}*{SmoothGrad} & sum & 5 &69.03&93.08&68.92&92.59&68.05&90.52&56.77&86.58&59.36&86.72\\
                                  & max & 5 &67.83&85.77&69.64&86.12&65.32&82.65&53.00&80.64&59.84&81.43\\
        \hline
        \hline
        Random & - & - &40.28&40.71&42.66&42.92&39.23&39.18&27.41&27.17&30.79&31.34\\
    \end{tabular}
    \end{adjustbox}
    \caption{HPD Results
    (\dag Note that evaluation with the default number of samples was not possible due to computational limitations (100 samples already took 7h and 14h for Classif and QA model on SQuAD))}
    \label{tab:res:hpd}
\end{table*}

\begin{table*}[h]
    \centering
    \begin{adjustbox}{width=\textwidth}
    \begin{tabular}{lll|rr|rr|rr|rr|rr}
        &&&\multicolumn{2}{c|}{SQuAD}&\multicolumn{2}{c|}{New Wiki}&\multicolumn{2}{c|}{NYT}&\multicolumn{2}{c|}{Reddit}&\multicolumn{2}{c}{Amazon}\\
        Interpreter & Aggregation & Samples & Classif & QA& Classif & QA& Classif & QA& Classif & QA& Classif & QA\\
        \hline
        \hline
        \multirow{4}*{LIME\textsuperscript{\dag}} & \multirow{2}*{sum} & 10 &39.23&38.20&41.82&40.66&36.98&37.26&25.89&22.84&27.87&27.08\\
                            &                    & 100 &39.30&40.91&42.30&43.96&39.41&39.38&32.90&46.71&27.42&32.52\\
                            & \multirow{2}*{max} & 10 &91.76&91.54&94.38&88.83&93.24&85.01&107.89&110.55&93.77&95.83\\
                            &                    & 100 &125.98&176.07&124.96&162.51&133.54&184.66&171.42&305.21&151.52&232.94\\
        \hline
        \multirow{4}*{SHAP\textsuperscript{\dag}} & \multirow{2}*{sum} & 10 &73.24&67.24&74.17&67.85&71.42&68.34&91.28&83.95&68.02&60.68\\
                            &                    & 100 &42.27&42.09&44.80&45.60&37.31&43.69&34.30&40.05&28.76&34.64\\
                            & \multirow{2}*{max} & 10 &99.16&102.31&97.42&101.44&97.10&92.66&130.98&127.01&99.87&102.16\\
                            &                    & 100 &107.51&137.77&102.01&132.57&94.43&132.64&149.95&240.47&135.62&207.62\\
        \hline
        \hline
        \multirow{2}*{Saliency} & sum & - &37.29&39.92&40.88&40.14&35.14&34.23&19.10&19.77&22.90&23.34\\
                                & max & - &35.58&38.20&39.75&39.81&34.08&36.35&19.15&20.04&22.17&23.78\\
        \hline
        \multirow{2}*{Integrated Gradients} & sum & 50 &37.28& 37.32& 39.16& 40.19& 33.30& 33.04& 18.96& 20.60& 22.50& 24.60\\
                                            & max & 50 &35.71& 34.99& 38.74& 38.09& 32.55& 32.80& 18.41& 20.57& 21.80& 23.69\\
        \hline
        \multirow{2}*{SmoothGrad} & sum & 5 &38.13&37.22&41.29&40.15&35.16&33.04&19.40&20.33&23.34&23.11\\
                                  & max & 5 &37.55&36.61&40.29&40.04&34.85&35.98&19.23&19.62&22.45&22.69\\
        \hline
        \hline
        Random & - & - &37.70&37.34&40.63&40.52&34.87&35.06&19.24&19.79&23.21&23.70\\
    \end{tabular}
    \end{adjustbox}
    \caption{SNR Results
    (\dag Note that evaluation with the default number of samples was not possible due to computational limitations (100 samples already took 7h and 14h for Classif and QA model on SQuAD)) Examples for which noise cannot be estimated are omitted.}
    \label{tab:res:snr}
\end{table*}

\end{document}